\pdfoutput=1

\documentclass[11pt]{article}

\usepackage{EMNLP2022}

\newenvironment{enumeratesquish}{\begin{list}{\addtocounter{enumi}{1}\labelenumi}{\setlength{\itemsep}{0em}\setlength{\labelwidth}{0.5em}\setlength{\leftmargin}{\labelwidth}\addtolength{\leftmargin}{\labelsep}}}{\end{list}\setcounter{enumi}{0}}
\usepackage{times}
\usepackage{latexsym}
\usepackage{graphicx}
\usepackage{float}
\usepackage{booktabs}
\usepackage{multirow}
\usepackage[labelformat=simple]{subcaption}

\graphicspath{ {./images/} }

\usepackage[T1]{fontenc}

\usepackage[utf8]{inputenc}

\usepackage{microtype}

\usepackage{inconsolata}

\usepackage{xspace}
\newcommand{\agree}{\textsc{AGReE}\xspace}

\title{\agree: A system for generating Automated Grammar Reading Exercises}

\author{Sophia Chan \\
        Educational Testing Service Canada \\
        \texttt{schan@etscanada.ca} \\
        \And  
        Swapna Somasundaran \\
        Educational Testing Service \\
        \texttt{ssomasundaran@ets.org} \\
        \AND
        Debanjan Ghosh \\
        Educational Testing Service \\
        \texttt{dghosh@ets.org} \\
        \And 
        Mengxuan Zhao \\
        Educational Testing Service Canada \\
        \texttt{mzhao@etscanada.ca}
        }

\begin{document}
\maketitle
\begin{abstract}
We describe the \agree system, which takes user-submitted passages as input and automatically generates grammar practice exercises that can be completed while reading. Multiple-choice practice items are generated for a variety of different grammar constructs: punctuation, articles, conjunctions, pronouns, prepositions, verbs, and nouns. We also conducted a large-scale human evaluation with around 4,500 multiple-choice practice items. We notice for 95\% of items, a majority of raters out of five were able to identify the correct answer and for 85\% of cases, raters agree that there is only one correct answer among the choices. Finally, the error analysis shows that raters made the most mistakes for punctuation and conjunctions.


\end{abstract}

\section{Introduction} \label{section:intro}
Acquiring a language necessitates learning its grammar.  In the United States, the Common Core standards for K-12 English literacy reflect this by including grammar as a learning outcome across all grade levels.\footnote{\url{http://www.corestandards.org/ELA-Literacy}} While both students and educators acknowledge that learning grammar is ``necessary and effective'' for language acquisition, it is ``not something they enjoy doing'' \citep{jean2011grammar}. Given the sheer amount of grammar constructs and rules, creating exercises that target each of these rules can be tedious. Likewise, for students, completing these exercises can become repetitive. 
To support teachers and more generally a growing interest in formative and computer-based assessments in the educational testing field, the demand for automatically created multiple-choice questions is  growing \citep{gierl2021advanced}. 





We introduce \agree: a system for generating \textsc{A}utomated \textsc{G}rammar \textsc{Re}ading \textsc{E}xercises that presents practice questions in a game-like setup (Figure \ref{grammar-question-frontend}).  While gamification is not an explicit part of our system, \agree offers immediate feedback, similar to a more formal incentive system that can boost user engagement and motivation by rewarding correct answers  \citep{plass2015foundations}. In related research \citet{yip2006online,hung2018scoping} find that games which focus on drilling and practice have a positive effect on language acquisition. The grammar questions are embedded within a text in the form of masked sentences, and users are encouraged to solve each question to uncover the whole text. Finally, \agree also allows you to attach any type of reading material for practice; giving learners agency over their choice of reading material has been shown to improve learner engagement \citep{moley2011moving}.


\begin{figure}
    \centering
    \includegraphics[width=8cm]{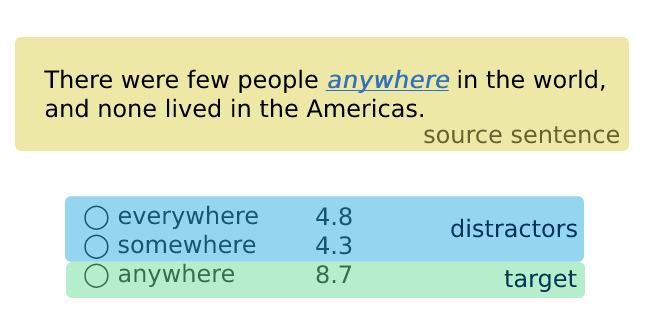}
    \caption{The components of a grammar question are: source sentence, target, and distractors. The contextual log probability score from BERT is shown to the right of each choice. In the actual task the target word is replaced by a blank.}
    \label{example}
\end{figure}

\begin{figure*}[t]
    \centering
    \frame{\includegraphics[width=15cm]{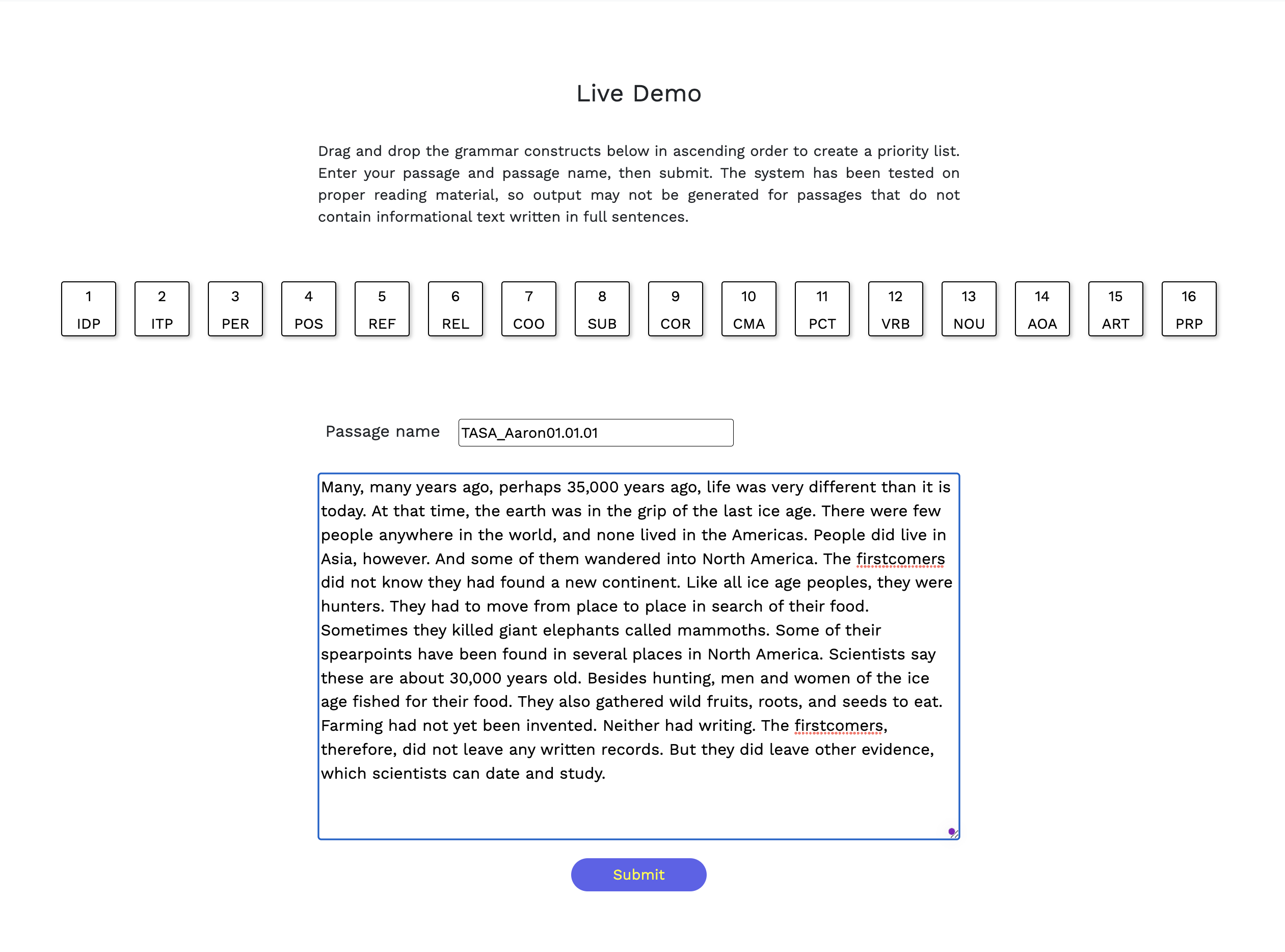}}
    \caption{The passage submission page accepts texts up to 1,000 characters long. Since \agree designs to generate items for only one construct per sentence, learners can re-order the grammar constructs as per their choice above the text box by dragging and dropping the numbered boxes.}
    \label{fig:submit-passage}
\end{figure*}

The generated questions in \agree are a variation on the Cloze task \citep{taylor1953cloze}. In this task, part of a sentence is removed and replaced with a blank and the goal is to recover the missing portion. While originally introduced to evaluate a text's readability, it has since been widely adapted for language practice and assessment. Figure \ref{example} illustrates an example of a practice item in \agree. In the running example in the paper, we use the source sentence ``There were \dots in the Americas''. The sentence contains the correct \textbf{target} word ``anywhere'' which has been replaced with a blank in the Cloze task. We use the pretrained language model BERT \cite{devlin2018bert} to generate the options for the blank via masked language modeling. BERT suggests the two \textbf{distractors} -- incorrect but plausible choice -- ``everywhere'' and ``somewhere''.


For \agree, we generate multiple-choice practice items for seven different constructs: \textit{punctuation}, \textit{article}, \textit{conjunctions}, \textit{pronouns}, \textit{preposition}, \textit{verb}, and \textit{noun}. Details and descriptions of these constructs are provided in Appendix \ref{sec:appendix}. Given a sentence, we identify the token corresponding to the above constructs and mask it from the sentence to form the question item. Next, we use BERT to identify the distractor choices (Section \ref{subsection:distract}). Note, we did not use any custom tokenizer here but instead used the default tokenizer from BERT.


After we design the grammar practice tool \agree, we conducted a thorough user-study via the crowdsourcing platform Amazon Mechanical Turk (MTurk). We ask the crowd-raters to attempt around 4,500 grammar items and then asked them to provide feedback on the quality of the items (see Section \ref{section:eval}). We notice that in 95\% of the items, a majority of crowd-raters were able to identify the correct answer whereas around 85\% of responses agree that the items have only one correct answer. On the contrary, in our error analysis we found that crowd-raters often made mistakes for punctuation and conjunction items showing these two constructs are harder than the rest. The grammar practice tool \agree is available online for practice and learning.\footnote{\url{https://grammarcloze.nlplab-dev.c.ets.org}}

\section{Overview} \label{section:over}

\begin{figure*}[t]
    \centering
    \frame{\includegraphics[width=15cm]{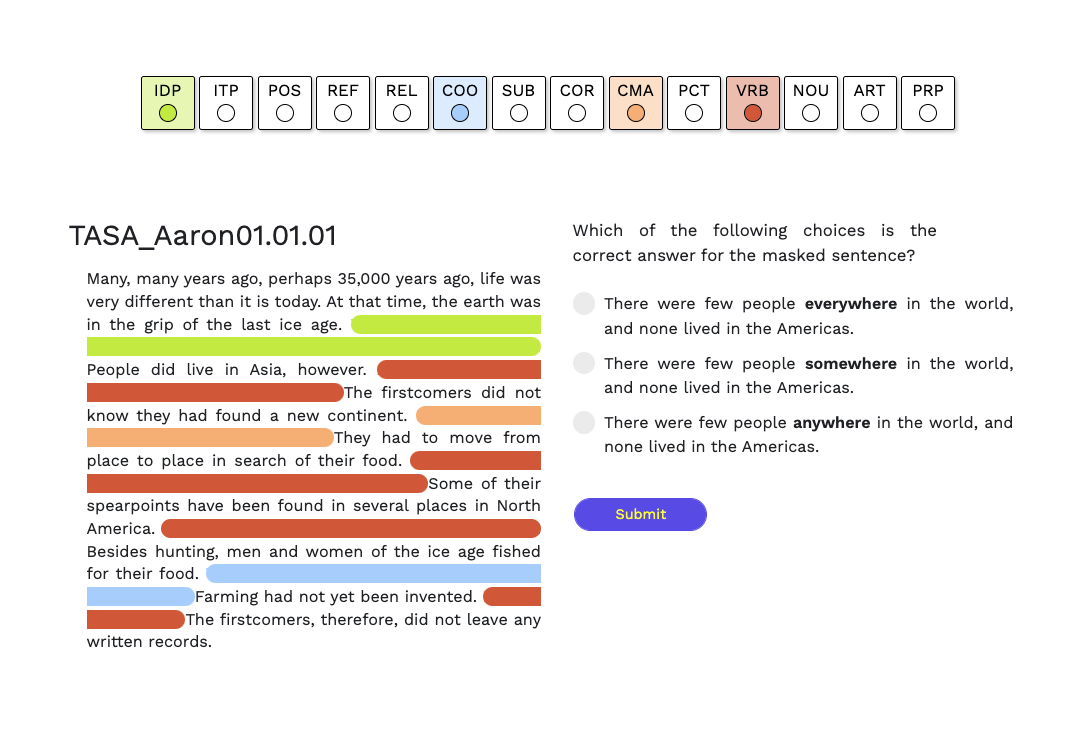}}
    \caption{After a user clicks on a masked sentence on the left side, a grammar question is revealed to the right of the window. The masked sentences are color-coded depending on the grammar constructs, and the short name for each construct can be found in the banner at the top of the page. See Appendix \ref{sec:appendix} for the full names of the grammar constructs, which also appear under the banner when hovered over. The passage shown is an excerpt from the TASA corpus we describe in Section \ref{section:over}.}
    \label{grammar-question-frontend}
\end{figure*}

To begin interacting with our tool, a user must submit a passage from the live demo page shown in Figure \ref{fig:submit-passage}. The available grammar constructs are shown above the free text box, and can be reordered in terms of priority from left to right. This enables the teacher or student to customize the experience based on learning goals. After the passage is submitted, the text is sent to a back-end implemented using Amazon Web Services (AWS) components.


After submission, the system will create grammar questions from the passage (see Figure \ref{grammar-question-frontend}). Grammar item generation happens at the sentence level. Since completing a Cloze task relies on filling in a blank based on the surrounding context \citep{taylor1953cloze}, we create an item out of every other sentence so that the user is provided with both enough context to fill in the blank, and with enough practice as they progress through their reading. Only one grammar item is generated per sentence, and this item will correspond to the first grammar construct that can be found in the sentence that our distractor generation algorithm successfully generates an item for.



Next, the grammar items  will appear to the user in the form of masked sentences. When a sentence is masked, it becomes clickable to the user. And once clicked on, the question choices will appear on the right side of the window. If the student selects an incorrect answer they are prompted to try again. But when the correct answer is selected the sentence becomes unmasked so that they may proceed with their reading. For each correct answer \agree provides feedback to encourage and motivate the learner.

The system has been tested thoroughly on informational texts such as the Touchstone Applied Science Associates (TASA) corpus which ``consists of representative random samples of text of all kinds read by students in each grade through first year of college'' \cite{ZenoEtAl1995EducatorsWordFrequency,landauer1998introduction}. Note that the system itself does not have any dependency to a particular corpus, and so in theory can be used with texts from any domain. 

\section{Grammar item generation} \label{section:gener}
Creating a grammar item from a sentence involves the following steps:
\begin{enumeratesquish}
    \item Token matching 
    \item Syntactic pattern matching 
    \item Sentence validation
    \item Sentence filtering
    \item Distractor generation
\end{enumeratesquish}

We use the following sentence ``There were few people anywhere in the world, and none lived in the Americas'' as our running example to show how to create the grammar item (Figure \ref{example}).

\subsection{Token matching}\label{token-matching}
Let's say the first grammar construct on the priority list is \textit{indefinite pronoun}. First, we determine whether the sentence contains a token in our pre-defined list of indefinite pronouns: \textit{everybody}, \textit{everywhere}, \textit{everything}, \textit{somebody}, \textit{somewhere}, \textit{something}, \textit{anybody}, \textit{anywhere}, \textit{anything}, \textit{nobody}, \textit{nowhere}, \textit{nothing}.\footnote{\url{https://www.gingersoftware.com/content/grammar-rules}}

\subsection{Syntactic pattern matching}
The system finds \textit{anywhere}, and this prompts a check to make sure the syntactic properties of the token match those of the grammar construct. Each grammar construct has a pattern defined using dependency parse tags and part of speech tags from spaCy.\footnote{\url{https://spacy.io}} 


\subsection{Sentence validation}
The syntactic properties of \textit{anywhere} matches the pattern defined for the indefinite pronoun construct, so at this point we have identified the grammar construct in the sentence and know its location. The token \textit{anywhere} is replaced with a \texttt{[MASK]} token, and we use the pre-trained language model BERT \citep{devlin2018bert} (as implemented in the Hugging Face \citep{wolf2019huggingface} repository)  to predict the most likely substitution for \texttt{[MASK]}. We currently only handle words that are common enough to exist in the vocabulary of the BERT tokenizer; our closed classes of distractor choices can all be found, but we may skip over an open class word if it turns out that the noun or verb is tokenized into separate wordpieces. The other tokens from the set of indefinite pronouns (Section \ref{token-matching}) become distractor candidates that feed into our distractor generation step in Section \ref{subsection:distract}.


\subsection{Sentence filtering}
If the word that was originally in the sentence (here, \textit{anywhere}) does not rank the highest in terms of probability score among the construct type (here, indefinite pronouns) as predicted by BERT in the previous step, it indicates that there is ambiguity in the correct answer to fill the \texttt{[MASK]} location with. In other words, multiple indefinite pronouns may suit the context. In our example \textit{anywhere} ranks the highest among the set of the indefinite pronouns at the \texttt{[MASK]} location, so we continue onto the next step for item generation. Otherwise, we discard the item from further processing. 

\subsection{Distractor generation} \label{subsection:distract}
It is important for a Cloze task that the distractors are plausible yet incorrect. 
\citet{gao2020distractor} argued that the BERT-based predictions for a masked word fits the requirements of a Cloze task perfectly. In other words, the most likely substitutions for a \texttt{[MASK]} token can be used as distractors. Thus, similar to \citet{gao2020distractor}, based on the probability scores of the indefinite pronouns, we select the highest ranking candidates (except the token \textit{anywhere})  as the distractor.

For nouns and verbs, the process remains the same except parts-of-speech tags are used in the place of a pre-defined list of tokens. We use the lemminflect package to create the list of possible candidates by inflecting the stem word.\footnote{\url{https://lemminflect.readthedocs.io}} 

For comma items, the candidates are all possible relocations of the comma. We create the distractor candidates by inserting a \texttt{[MASK]} token in-between every word. 


\section{Evaluation} \label{section:eval}

\begin{figure}[t]
    \centering
    \includegraphics[width=6cm]{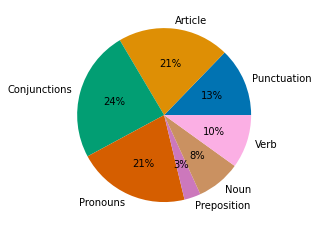}
    \caption{The distribution of grammar constructs in generated questions presented to Turkers (N=4,568).}
    \label{fig:all-pie-chart}
\end{figure}

\begin{figure}[t]
    \centering
    \includegraphics[width=6cm]{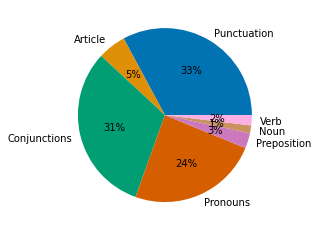}
    \caption{The distribution of grammar constructs in generated questions with two or fewer correct answers (N=455).}
    \label{fig:minority-pie-chart}
\end{figure}

We are interested in evaluating the quality of the multiple-choice items in \agree. To that end, we conduct a large-scale MTurk study where we present crowd-raters with our generated grammar exercises as shown in Figure \ref{grammar-question-frontend}. We ask the raters to select their choice for each item, and to answer followup questions along four aspects of item quality. We paid raters \$1.70 USD for completing our prerequisite task and \$2.00 USD for completing our main task, which is approximately equivalent to \$20-\$24 USD (per hour). In total, the evaluation cost \$12,575 USD.

\subsection{Prerequisite task}
We conducted a set of prerequisite tasks to select the raters for our main task on \agree. A rater must have at least 10,000 approved HITs, $\ge$ 95\% of HIT approval, a Masters qualification, and reside  in the  United States. We also ask the raters to answer a set of 20 publicly available TOEFL Junior reading comprehension questions (typically used for assessing English language skills of students 11 or older) so that we can gauge raters' English proficiency.\footnote{\url{https://www.ets.org/toefl\_junior/prepare/standard\_sample\_questions/reading\_comprehension}} 

The mean score on this set of reading comprehension questions among 79 raters is 96\%, and 95\% of raters received a score of 90\% or higher, meaning they got at most two questions out of 20 incorrect. Responses from this high-performing group make up 99.8\% of our responses. Given that the raters have some competence in the language, we expect that they will be able to identify the correct answer among the choices. If they are unable to do so, it will provide a strong indication that our generated items are incorrectly keyed.


\subsection{Main task}

\label{sec:main-task}

Raters are presented with two passages containing grammar questions from \agree. The setup is identical to how we present the questions on the front-end  (Figure \ref{grammar-question-frontend}), and the method for generating grammar items is the same as described previously (Section \ref{section:gener}). Each passage is around 10 sentences long, taken from the beginning of either a TASA \citep{ZenoEtAl1995EducatorsWordFrequency} or a TIPSTER \citep{harman1993overview} document. Altogether 4,568 grammar constructs items were presented to the raters. Five different raters responded to each item to select the correct answer. The distribution of the items is shown in Figure \ref{fig:all-pie-chart}.




Among the 4,568 items that we generated, for 76\% of items all five raters answered correctly. If we consider items where the majority of raters answered correctly (i.e, three or more selected the target), then the proportion increases to 95\%. This is an indication that nearly all our items are correctly keyed. On the contrary, there were 455 instances where only a minority of raters chose the correct answer. Among these items, the article, verb, and noun items are less frequent whereas the punctuation and conjunctions items are the most frequent (see Figure \ref{fig:minority-pie-chart}).

To shed some light on why raters made mistakes on certain items, we randomly select one example each of punctuation and conjunctions items where no raters answered correctly. Note that we present the choices here in order of their ranking from BERT (the log probability is in parentheses), but when presented to raters the choices are randomly shuffled. Here is an example of a problematic punctuation (comma) item:

\begin{quote}
    \textbf{source sentence}: With so many children in the family\_\_\_
    there\_\_\_ was a constant\_\_\_ buzz of activity \\
    \textbf{target}: ... family\textbf{,} ... (12.76) \\
    \textbf{distractor}: ... constant\textbf{,} ... (7.17) \\
    \textbf{distractor}: ... there\textbf{,} ... (6.07)
\end{quote}

And here is an example of a problematic conjunctions (coordinating conjunction) item:
\begin{quote}
    \textbf{source sentence}:  Many of these people did not go to the theater, of course, \_\_\_ to keep playgoers happy, acting troupes had to provide a variety of plays. \\
    \textbf{target}: but (13.93) \\
    \textbf{distractor}: and (13.39) \\
    \textbf{distractor}: so (12.18)
\end{quote}

In both items, all five raters selected the highest-ranking distractor as their answer. This supports the idea that the contextual probabilities are useful predictors of distractor-context fit. While the placement of the comma in the target is the only proper usage in the punctuation example, one might argue that both \textit{but} and \textit{and} are grammatical in the conjunctions source sentence. However, the use of the connective \textit{but} more accurately describes the relationship between the first and second clause. Given that the probability of the distractor and target are so close in the conjunctions example, we suspect that the distance between the probabilities provides some signal about item difficulty, but future work is needed to investigate and calibrate the difficulty of the grammar items.

\begin{figure*}[t]
    \begin{subfigure}[t][][t]{.233\textwidth}
        \centering
        \includegraphics[width=\textwidth]{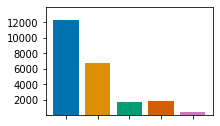}
        \caption{\textit{I felt that there was more than one correct answer}.\\ }
        \label{subfig:more-correct}
    \end{subfigure}
    \hfill
    \begin{subfigure}[t][][t]{.196\textwidth}
        \centering
        \includegraphics[width=\textwidth]{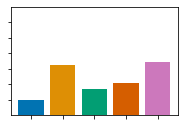}
        \caption{\textit{I felt that some response options were too obviously wrong}.}
        \label{subfig:obviously-wrong}
    \end{subfigure}
    \hfill
    \begin{subfigure}[t][][t]{.196\textwidth}
        \centering
        \includegraphics[width=\textwidth]{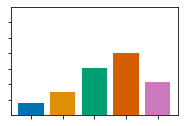}
        \caption{\textit{The question was likely created by a teacher}.}
        \label{subfig:teacher}
    \end{subfigure}
    \hfill
    \begin{subfigure}[t][][t]{.303\textwidth}
        \centering
        \includegraphics[width=\textwidth]{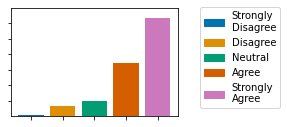}
        \caption{\textit{The question was easy}.\\ }
        \label{subfig:easy}
    \end{subfigure}
    \hfill
    
    \caption{Bar charts displaying the distribution of responses for each item quality aspect over the five Likert scale options (N=22,840).}
    \label{fig:bar-charts}
\end{figure*}

\



The purpose of our MTurk experiment is not only to measure how many items are solvable by the raters, but also to know specific aspects about them. Such as whether (1) the items are correctly keyed (i.e., we have accurately and unambiguously identified the target) (2) the item contains nonfunctional distractors \citep{gierl2017developing}, (3) the item can be distinguished from a human-generated one, and finally, (4) the item was difficult. To answer these four questions we also asked the same group of raters to respond a five-point Likert scale (Strongly Disagree, Disagree, Neutral, Agree, Strongly Agree) to the following statements, respectively.

\begin{enumerate}
    \item \textit{I felt that there was more than one correct answer}
    \item \textit{I felt that some response options were too obviously wrong}
    \item \textit{The question was likely created by a teacher}
    \item \textit{The question was easy}
\end{enumerate}



Figure \ref{fig:bar-charts} presents four bar charts displaying the distribution of the responses to the above four questions from left to right. We find corroborating evidence that the items are correctly keyed: 85\% of responses agree that the items have only one correct answer (Figure \ref{subfig:more-correct}). As for whether some options are too obviously wrong, the results appear to be mixed; no clear pattern in the responses can be observed (Figure \ref{subfig:obviously-wrong}). We find that more often than not, raters found our items indistinguishable from those created by teachers, but here we also do not observe a clear pattern (Figure \ref{subfig:teacher}). Finally, nearly all responses say that the items are easy. Gathering data from second language learners could help clarify whether the items are easy for English learners as well (Figure \ref{subfig:easy}).






Overall, the MTurk results paint a promising picture for the utility of our items. Since 95\% of items were answered correctly by a majority of raters, and 85\% agree that there was only one correct answer, we have strong indicators that nearly all our items are correctly keyed.



 
\section{Related work}

Tools that enhance authentic texts in support of grammar acquisition include WERTi \citep{meurers2010enhancing}, FLAIR \citep{chinkina2016linguistically}, and GrammarTagger \citep{hagiwara2021grammartagger}. Except for a verb practice activity in WERTi, these tools 
provide few opportunities for immediate feedback. Our system fills this gap by generating multiple-choice practice questions from authentic texts.

On the other hand, there are also systems that generate multiple-choice grammar questions. FAST \citep{chen2006fast} covers nine grammatical categories and \citet{lee2016call} create a system for learning preposition usage. We extend these works by not only covering a range of grammatical categories and collecting perception responses, but also conducting a large-scale evaluation on item performance (i.e., whether the item can be solved by the user with some competence in the language).

\section{Conclusion} 
We describe \agree, a system and procedure for converting an informational passage into game-like grammar practice exercises that can be completed while reading. We find in human evaluations that nearly all the multiple-choice questions we generate for the exercises are correctly keyed, and can therefore be used to provide immediate feedback to students. We also observe for almost 95\% of items that a majority of the raters were able to identify the correct target. On the contrary,  raters made the most mistakes for punctuation and conjunctions. Although we did not design our system to include gamification explicitly, it is set up in a way that makes it easy to incorporate in the future. The system can automatically generate a wealth of questions for which the correct answer is identified, and these questions can be used by whomever (e.g., game designers) to create a game that rewards learners when they correctly solve them. 


\section{Future work}
\agree is a proof of concept system we built with the intent of allowing users to personalize grammar items according to their unique goals. The user interface does not currently prevent learners from progressing through the passage without providing answer for each question. In fact, it is likely that they can still complete the reading since it is usually a weakly semantic element missing from the sentence. A truly gamified system may choose to block out the rest of the passage before the current question is answered, or make it so that the next question cannot be clicked on until the current question is answered. 

As it stands, the system allows teachers some control over the generated items; the reading material itself and the ranking of grammar constructs can be customized. However, teachers do not have control at the level of individual items. We may want to build in this finer-grained control in the future so that the exercises can be adapted more closely to their needs. Currently, the number of questions generated for a given construct (e.g., preposition) depends on that construct's ranking, its frequency in the text, and how well it is covered in the manually created list of distractor choices. For example, there may be fewer preposition items generated than expected due to the fact that we use a reduced set of prepositions in \agree. If we expand the set of distractor choices to cover all possible prepositions, we would likely run into latency issues.

While increasing the coverage of existing constructs is one potential line of future work, it may be more important to find ways to align our constructs to existing EFL (English as a Foreign Language) curricula if we want to create efficacious questions. Since the constructs that a learner struggles with is influenced by\textemdash among other factors\textemdash aspects of their learner profile such as English level \citep{hawkins2010criterial} and language background, an efficacious system should take these aspects into account. Doing so may allow \agree to provide more personalized feedback and generate distractors that can be calibrated. For example, we can imagine using token-level probabilities to make filtering decisions about whether an item is appropriate for a certain language level. The distance between the target log probability and the log probability of the hardest distractor can be smaller for a learner whose language level is higher, assuming that as language ability improves, so does the ability to discriminate between a correct choice and a plausible yet incorrect one.

This raises the question about the threshold at which a token can be considered grammatical versus not, as illustrated in the coordinating conjunction example in Section \ref{sec:main-task}. According to \citet{larsen2001teaching}, one way to think about grammar is to see it as an interaction between form/structure, meaning/semantics, and use/pragmatics. \citet{schneider2016detecting} also argue that, when it comes to learner English, there is no ``clear dichotomy between innovation and error''. Following this line of thinking, we view the threshold for grammaticality as context-specific and tied to pedagogical goals. 

More experimentation is also needed to determine how the quality of the questions changes when individual components are altered or replaced. To increase the flexibility of the system, we can think of replacing the token matching and syntactic pattern matching components with a more specialized model that identifies the gap locations, such as the one described in \citet{felice2022constructing}. As such, making the system more functional would involve building in the ability to evaluate the output of individual components in addition to the final output.

\newpage
\bibliographystyle{acl_natbib}
\bibliography{anthology,custom}

\appendix
\section{Grammar constructs}\label{sec:appendix}
\begin{table*}
    \centering
    \begin{tabular}{rrl|l}
        \hline
         Construct & &  & Distractor candidates  \\
         \hline
         \multirow{2}{*}{Punctuation} & Comma location & CMA & All possible relocations \\
         & Punctuation  & PCT & : : , \\
         \hline 
         Article & & ART & the, a, an \\
         \hline
         \multirow{5}{*}{Conjunctions} & Coordinating conjunction & COO & for, nor, but, or, yet, so, and \\
         & Subordinating conjunction & SUB & after, although, because, before, if, once, since, \\
         & & & than, unless, until, when, whenever, while, as \\ 
         & Correlative conjunction & COR & either/or, neither/nor, both/and, as/so, \\
         & & &  whether/or \\ 
        \hline 
        \multirow{7}{*}{Pronouns} & Indefinite pronoun & IDP & everybody, everywhere, everything, somebody, \\
        & & & somewhere, somewhere, something, anybody, \\
        & & & anywhere, anything, nobody, nowhere, nothing \\
        & Interrogative pronoun & ITP & who, which, what, whose, whom \\
        & Possessive pronoun & POS & my, mine, your, yours, our, ours, their, theirs \\
        & Reflexive pronoun & REL & myself, yourself, herself, himself, itself, \\
        & & & yourselves, ourselves, themselves \\
        \hline 
        Preposition & & PRP & to, toward, on, onto, in, into \\
        \hline
        Noun & & NOU & NN, NNS \\
        \hline
        Verb & & VRB & VB, VBD, VBG, VBN, VBP, VBZ \\
        \hline
        
    \end{tabular}
    \caption{The available grammar constructs and their distractor candidates. All the possible tokens are enumerated except for \textit{comma location} where the candidates are all possible relocations of the comma, and \textit{noun} and \textit{verb} items where candidates are inflections of the word stem.}
    \label{tab:grammar-constructs}
\end{table*}
Each grammar construct and its sub-constructs are listed in Table \ref{tab:grammar-constructs}, along with the set of distractor candidates used for the token matching step from Section \ref{token-matching} and distractor candidate generation step from Section \ref{subsection:distract}.

\end{document}